\documentclass[10pt,twocolumn,letterpaper]{article}

\usepackage[pagenumbers]{cvpr} 

\usepackage{booktabs,multirow,makecell,array,siunitx}
\newcolumntype{L}[1]{>{\raggedright\arraybackslash}p{#1}}

\usepackage{siunitx}

\usepackage{pifont}
\newcommand{\cmark}{\ding{51}} 
\newcommand{\xmark}{\ding{55}} 

\definecolor{cvprblue}{rgb}{0.21,0.49,0.74}

\usepackage{capt-of}  

\usepackage[pagebackref,breaklinks,colorlinks,allcolors=cvprblue]{hyperref}



\def\paperID{} 
\def\confName{CVPR}
\def\confYear{2026}

\title{LinkedOut: Linking World Knowledge Representation Out of Video LLM 
\\ for Next-Generation Video Recommendation}

\author{
Haichao Zhang$^{1,\dagger,\ddagger}$ \quad
Yao Lu$^{2,\dagger}$ \quad
Lichen Wang$^{2,\dagger}$ \quad
Yunzhe Li$^{2}$ \quad \\
Daiwei Chen$^{3}$ \quad
Yunpeng Xu$^{2}$ \quad
Yun Fu$^{1}$ \\
$^{1}$Northeastern University \quad
$^{2}$LinkedIn \quad
$^{3}$University of Wisconsin--Madison \\
}

\begin{document}

\maketitle

\begingroup
\renewcommand{\thefootnote}{\fnsymbol{footnote}}
\footnotetext[2]{Equal Contribution.} 
\footnotetext[3]{Work done as an intern at LinkedIn (a Microsoft company).} 
\endgroup


\begin{abstract}
Video Large Language Models (VLLMs) unlock world-knowledge-aware video understanding through pretraining on internet-scale data and have already shown promise on tasks such as movie analysis and video question answering. However, deploying VLLMs for downstream tasks such as video recommendation remains challenging, since real systems require multi-video inputs, lightweight backbones, low-latency sequential inference, and rapid response. In practice, (1) decode-only generation yields high latency for sequential inference, (2) typical interfaces do not support multi-video inputs, and (3) constraining outputs to language discards fine-grained visual details that matter for downstream vision tasks. We argue that these limitations stem from the absence of a representation that preserves pixel-level detail while leveraging world knowledge. We present \textit{LinkedOut}, a representation that extracts VLLM world knowledge directly from video to enable fast inference, supports multi-video histories, and removes the language bottleneck. \textit{LinkedOut} extracts semantically grounded, knowledge-aware tokens from raw frames using VLLMs, guided by promptable queries and optional auxiliary modalities. We introduce a cross-layer knowledge fusion MoE that selects the appropriate level of abstraction from the rich VLLM features, enabling personalized, interpretable, and low-latency recommendation. To our knowledge, \textit{LinkedOut} is the first VLLM-based video recommendation method that operates on raw frames without handcrafted labels, achieving state-of-the-art results on standard benchmarks. Interpretability studies and ablations confirm the benefits of layer diversity and layer-wise fusion, pointing to a practical path that fully leverages VLLM world-knowledge priors and visual reasoning for downstream vision tasks such as recommendation.
\end{abstract}

\begin{figure}[t]
\centering
\includegraphics[width=0.95\linewidth]{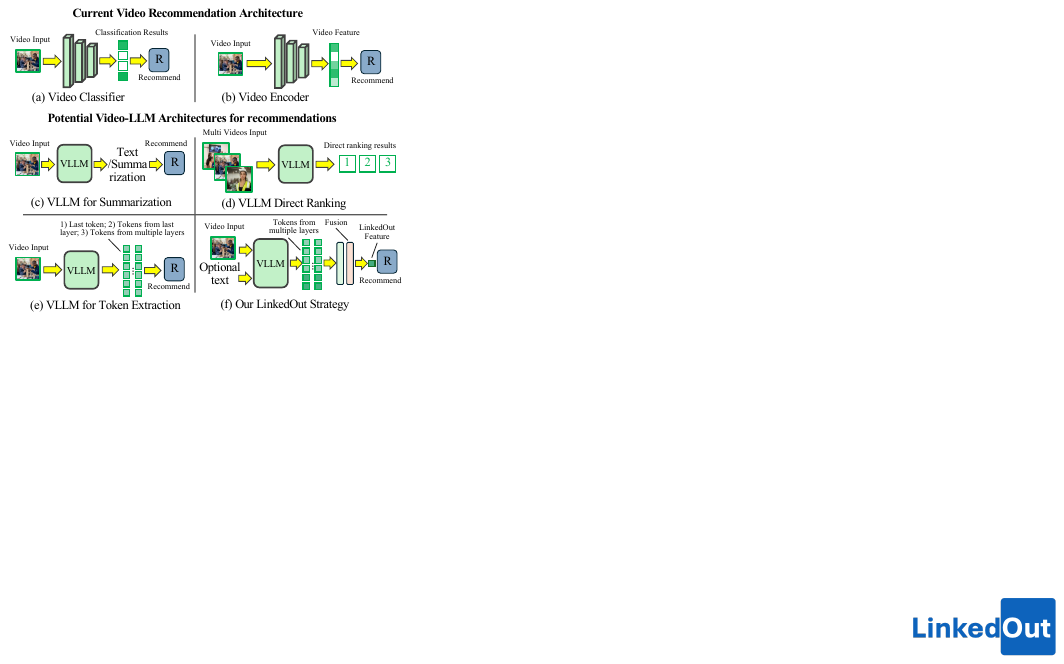}
\caption{Conceptual differences of our LinkedOut and other methods. Existing approaches using video models either as a classifier or a feature extractor (i.e., (a) and (b)) which are trained on task specific data without LLM/world knowledge. Other solutions such as (c) VLLM for content summarization, (d) VLLM direct video ranking, or (e) VLLM for token level feature extraction, they can neither fully utilize the world knowledge from VLLM nor are computationally feasible for recommendation online serving. Our LinkedOut approach is able to fully utilize both pixel level visual information with VLLM knowledge. Videos are eventually transferred to feature vectors for fast online recommendation serving.}
\label{fig:framework_concept}
\vspace{-2mm}
\end{figure}

\section{Introduction}\label{sec_intro}
Video recommendation is the primary interface through which viewers discover content at scale~\cite{covington2016deep}. Its inherent need for lightweight backbones, low-latency sequential inference, and rapid response makes it a representative downstream setting for studying foundation-model integration. 
Despite remarkable advances in representation learning and ranking architectures \cite{rendle2010factorization, kang2018sasrec, sun2019bert4rec, naumov2019dlrm, zhou2018din}, production systems still depend heavily on hand-crafted or pre-extracted labels (e.g., categories, tags, reactions) distilled from raw videos. This label-centric design reduces input dimensionality and eases serving, but it also discards the majority of the information present in pixels, limits semantic coverage to what was annotated in the training set, and makes personalization brittle in cold-start and long-tail regimes. Moreover, pipelines that summarize videos into text before recommendation inherit a language bottleneck: they constrain the system to preselected features and textual abstractions rather than the full visual token space. In particular, such pipelines struggle to recognize nuanced attributes (e.g., visual humor and narrative pacing) or to leverage broader factual or commonsense context and to adapt quickly to fast-evolving trends without expensive retraining.

Multimodal LLMs have recently demonstrated strong transferability from internet-scale pretraining, aligning visual tokens with natural language and encoding broad world knowledge~\cite{radford2021clip, alayrac2022flamingo, li2023blip2, liu2023llava}. In video, joint training on raw frames and transcripts unlocks temporally grounded semantics and cross-modal reasoning~\cite{bain2021frozenintime, miech2019howto100m}. These models are promptable, so natural language can steer them to extract task-specific attributes and abstractions on demand without modifying the backbone. Transformer layers also specialize by depth: later layers aggregate global context and higher-level semantics, while earlier layers emphasize local patterns. This suggests an opportunity to link world knowledge and pixel-level cues from a video LLM directly to a recommender, without first reducing videos to text summaries, while adaptively selecting the semantic granularity that best serves each ranking decision. Adapting VLLMs to video recommendation is therefore a promising way to move beyond hand-crafted or pre-extracted label-centric pipelines and to bring world-knowledge-aware visual understanding and possible reasoning into recommendation.

Although many recent efforts have built video large language models (VLLMs) with world knowledge and reasoning that benefit tasks such as video content understanding, visual question answering, and video world modeling, the scope of these downstream uses remains narrow. They mainly target settings that can tolerate the current limitations of VLLMs. When a task requires multi-video inputs, lightweight backbones, low-latency sequential inference, and rapid response, these limitations become impractical for straightforward pipelines like Fig.~\ref{fig:framework_concept}(c–e).

Video recommendation is therefore a representative task for analyzing this problem.
 It must keep the pretrained world knowledge inside the VLLM, but its outputs are item indices from a database rather than natural language, so they do not follow the language distribution. Current VLLMs typically rely on text outputs due to the language interface. Some tasks, such as robotics planning, fine-tune VLLMs to produce structured actions, but this often narrows the output distribution and can lead to catastrophic forgetting of the original knowledge.
In addition, current VLLMs (and many LLMs) are slow at inference time because of the decode-only transformer architecture: every newly generated token must be fed back as input for the next step, which lengthens the reasoning process and is unsuitable for real-time video recommendation. Typical VLLM interfaces also lack native support for multi-video inputs. A single video can already occupy tens of thousands of tokens, and large token budgets further increase latency and computation, making it difficult to design or post-train VLLMs that accept multiple videos at once. However, video recommendation usually needs to consume a sequence of historical videos per user and to encode a large candidate pool. These challenges have so far limited the use of VLLMs in tasks that simultaneously require multi-video inputs, lightweight models, low-latency sequential inference, and rapid response, while still needing world knowledge for visual reasoning, content understanding, and generalization.

\begin{figure*}[t]
\centering
\includegraphics[width=0.98\linewidth]{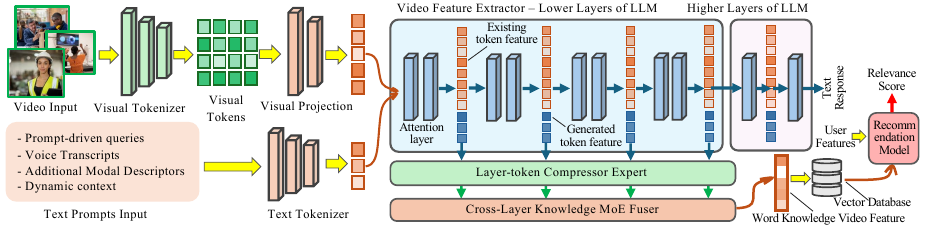}
\caption{LinkedOut framework. Raw video frames are tokenized and projected into a pretrained video LLM. The optional text queries (e.g. transcripts, metadata, and dynamic context) are co-tokenized and injected to guide extraction. As tokens traverse lower to higher transformer layers, cross-modal attention enriches the sequence with both existing visual/text tokens and LLM-generated knowledge tokens. A Layer-token Compressor Expert condenses token sequences, and a Layer Knowledge Fuser MoE gates across layers to adaptively select the appropriate semantic depth, yielding a unified, world-knowledge-aware video embedding. At serving time, user features and video embeddings are passed to a lightweight recommendation model to produce a relevance score. The promptable design supports flexible modality fusion and selective fine-tuning, enabling personalized, interpretable, and cold-start-robust video recommendation.}\label{fig:framework}
\vspace{-2mm}
\end{figure*}

To this end, we introduce \textbf{\textit{LinkedOut}}, a knowledge-aware, modality-extendable video recommendation framework that links world knowledge out of video pixels through a video LLM. The general structure is shown in Figure~\ref{fig:framework}. To address the language-output constraint, we do not fine-tune the VLLM into the recommendation space, which would risk domain shift and catastrophic forgetting. Instead, we directly extract embeddings from intermediate token representations of the VLLM, since world knowledge is distributed across transformer layers and is computed jointly with the input tokens. To reconcile the different abstraction levels encoded across the backbone, we propose a cross-layer Knowledge-fusion Mixture-of-Experts that fuses representations from multiple transformer blocks. Concretely, \textit{LinkedOut} combines both existing tokens and generated tokens within each layer, then applies cross-layer gating to produce a unified recommendation embedding that blends fine-grained visual cues with high-level conceptual knowledge~\cite{fedus2021switch, riquelme2021moe}.

To address slow inference and enable scalable deployment, \textit{LinkedOut} adopts a store-and-retrieve architecture. An offline pipeline first precomputes compact \textit{LinkedOut} features with the Cross-layer Knowledge-fusion MoE and stores them in an embedding database. At serving time, a lightweight recommendation module retrieves candidate embeddings and produces fast rankings conditioned on user context and prompts. This design preserves the benefits of content-aware modeling while keeping latency low, and it allows new modalities to be added offline without retraining the online ranker. It also strengthens support for multi-video inputs, since feature extraction and heavy reasoning are isolated from the online VLLM path, avoiding the quadratic token cost of feeding multiple videos into the VLLM at serving time.
To move beyond traditional video recommendation that relies on tags or preselected features, our approach replaces fragile label taxonomies with dense, semantically grounded token embeddings extracted directly from raw frames. A promptable feature-focus module steers the backbone toward user-relevant attributes or auxiliary modalities (for example, transcripts, audio, metadata), or toward dynamic context (for example, trending topics) using text prompts, which enables rapid adaptation without retraining.
Empirically, \textit{LinkedOut} achieves state-of-the-art performance on public video recommendation benchmarks, benefiting from VLLM world knowledge aggregated across layers.
\noindent Our main contributions are as follows:
\begin{itemize}
\item We propose \textit{LinkedOut}, a knowledge-aware, modality-extendable video recommendation framework that links world-knowledge features out of video pixels via a video LLM, thereby removing the language bottleneck. To the best of our knowledge, this is the first VLLM-based recommendation system that integrates a video LLM and consumes raw video frames directly, while avoiding text-only summarization bottlenecks and fully leveraging web-scale factual and commonsense knowledge from pretrained VLLMs.

\item We design a \textit{Cross-layer Knowledge-fusion MoE} that selects and concentrates the appropriate abstraction from different depths of intermediate VLLM tokens, producing a unified embedding that mixes fine-grained visual cues with high-level semantics under strict latency constraints, outperforming last-layer or last-token baselines and improving robustness across genres and user intents.

\item For fast inference, we use a store-and-retrieve pipeline that precomputes compact \textit{LinkedOut} features offline and supports low-latency online ranking with multi-video histories. On public video recommendation benchmarks, \textit{LinkedOut} achieves state-of-the-art results, benefiting from world-knowledge priors concentrated across multiple layers of the VLLM.
\end{itemize}

\section{Related Work}\label{sec_related}
\subsection{Multimodal LLMs for Video Understanding}
Web-scale foundation models have substantially advanced multimodal representation learning by aligning visual inputs with natural language supervision.
Early work such as CLIP learns image--text alignment via contrastive pretraining and enables strong zero-shot transfer across recognition tasks \cite{radford2021clip}.
Building on this paradigm, Flamingo combines frozen vision encoders with a causal language model through gated cross-attention to support in-context learning across a broad set of vision--language problems \cite{alayrac2022flamingo}.
BLIP-2 introduces a lightweight query transformer to connect pretrained visual encoders with LLMs, improving data efficiency and generalization \cite{li2023blip2}, while LLaVA scales visual instruction tuning to align LLMs with image-grounded instructions and open-ended dialogues \cite{liu2023llava}.

Extending multimodal LLMs from images to video requires modeling temporal dynamics and long-range dependencies.
Large-scale narrated video corpora such as HowTo100M provide web-scale paired video--text supervision that has catalyzed progress in video--language representation learning \cite{miech2019howto100m}.
Architectures such as Frozen-in-Time jointly encode frames and text to support retrieval and transfer, offering an effective foundation for downstream video understanding \cite{bain2021frozenintime}.
More recent video LLM systems further combine frame-wise attention with instruction tuning to elicit temporally grounded semantics and world knowledge from long videos \cite{alayrac2022flamingo,li2023blip2}.
In parallel, efficiency and evaluation for video understanding remain central challenges: VQToken proposes extreme token reduction to enable lightweight video-LLM inference with a minimal token budget \cite{zhang2025vqtoken}, and Dense Video Understanding introduces dense, fine-grained evaluation settings that stress temporally detailed multimodal reasoning \cite{zhang2025dive}.

A consistent empirical trend in transformer models is that deeper layers aggregate broader context and encode higher-level abstractions, while earlier layers preserve more local and fine-grained patterns.
This motivates layer-aware representations that can match the abstraction level required by a downstream task.
Mixture-of-Experts (MoE) provides a scalable mechanism to route inputs to specialized experts and increase model capacity without a proportional increase in compute \cite{riquelme2021moe}, complementing MoE advances in language models such as Switch Transformers \cite{fedus2021switch}.
Our work leverages these insights by extracting layer-wise token representations from a video LLM and fusing them via a cross-layer MoE, producing a unified embedding that integrates fine-grained visual cues with higher-level, knowledge-grounded semantics.
Unlike prior MoE approaches that route across separate modules or models, we explicitly fuse cross-layer token features within a single VLLM, as different layers encode distinct semantic depths.

\subsection{Multimodal LLMs in Recommendation}
Traditional recommender systems rely on collaborative filtering \cite{rendle2010factorization, he2020lightgcn, guo2017deepfm} and sequential modeling \cite{hidasi2016gru4rec, kang2018sasrec, yuan2019nextitnet}, which operate primarily on item IDs. To address cold-start and long-tail limitations, content-aware recommendation leverages auxiliary modalities. Early approaches extract fixed labels or shallow features from pretrained networks \cite{covington2016youtube}. Recent multimodal recommenders integrate richer features through graph convolution (MMGCN \cite{MMGCN}) or graph-based representations (GRCN \cite{wei2020graph}). Recent foundation models offer a transformative alternative: CLIP-like features \cite{radford2021clip} enhance retrieval via zero-shot transfer, while multimodal LLMs such as Flamingo \cite{alayrac2022flamingo}, BLIP-2 \cite{li2023blip2}, and LLaVA \cite{liu2023llava} unlock flexible conditioning via natural-language prompts and provide web-scale world knowledge. For video recommendation specifically, three paradigms have been explored: first, using frozen pretrained video models to extract fixed features \cite{ni2023microlens}; second, jointly training video encoders end-to-end with recommendation objectives \cite{ni2023microlens}; and third, summarizing videos into text via captioning or ASR, then applying text-based LLMs. 


However, existing approaches typically suffer from three limitations: (1) frozen encoders treat multimodal features as static descriptors, limiting adaptability to recommendation-specific semantics and user context; (2) end-to-end training achieves better performance but incurs prohibitive computational costs; and (3) text-based summarization introduces a language bottleneck that discards pixel-level visual information while relying on hand-crafted prompts or fixed layers that may provide either insufficient or excessive abstraction. LinkedOut addresses these gaps by combining knowledge-aware extraction directly from video pixels, adaptive layer-wise fusion that selects appropriate semantic granularity per instance, and efficient decoupling of offline embedding computation from online inference, thereby bridging foundation model capabilities with production-grade recommender constraints.

\section{LinkedOut Approach}\label{sec_method}
\subsection{Preliminaries and Interpretability Analysis}
\textbf{Preliminaries on VLLMs.}
A typical video LLM (VLLM) first tokenizes each input modality (for example, video frames, text prompts, audio) with its corresponding tokenizer. The resulting tokens are projected to a common embedding space and concatenated to form a single token sequence, which is then fed into the LLM to perform reasoning with the world knowledge stored in the model \cite{storm2025token, streaming2024long, survey2023video, survey2025mllm}. The LLM itself is composed of multiple transformer layers. World knowledge is distributed across these pretrained transformer layers. At each layer, token representations are updated through self-attention and feed-forward blocks, producing new token embeddings that are passed to the next layer in an autoregressive next-token prediction manner.

\begin{figure*}[t]
\centering
\includegraphics[width=0.95\linewidth]{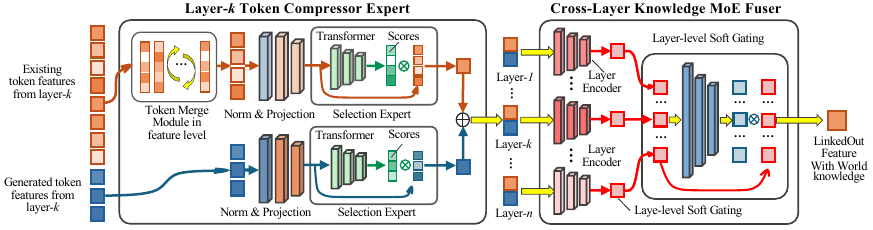}
\caption{Overview of the Cross-Layer Knowledge-Fusion MoE. \textit{Token Compressor Expert} receives existing and generated token features. A token-merge module aggregates redundant signals at the feature level, followed by normalization and projection. A lightweight selection expert predicts scores to reweight tokens; the two streams are then merged to yield a compact, semantically faithful representation for layer \textit{k}. \textit{Cross-Layer Knowledge MoE Fuser} treats each layer as an expert. Layer encoders summarize the compressed tokens, and layer-level soft-gating produces context-conditioned mixture weights that adaptively combine information across layers. The resulting LinkedOut feature fuses fine-grained visual details with world knowledge, providing a single, retrieval-ready embedding that improves efficiency while preserving recommendation-relevant semantics.}
\label{fig:framework_moe}
\vspace{-2mm}
\end{figure*}



\noindent\textbf{Old and new tokens.}
During decoding, each transformer layer first applies (multi-head) self-attention or cross-attention over all input tokens from all modalities. After this first pass, the model starts to generate tokens autoregressively: each newly predicted token is appended to the previous tokens and used as input for the next step. KV cache \cite{kvcache} can speed up this process, but the overall decoding is still sequential. For clarity, at a given layer we refer to tokens whose indices are within the original input length as \textit{old} tokens, and to tokens whose indices exceed the original input length as \textit{new} tokens.

\noindent\textbf{Layer-wise token representation and interpretability.}
Recent studies suggest that different transformer layers tend to capture different levels of knowledge~\cite{lepori2024beyond, raghu2021vision, liu-etal-2019-linguistic, chen2025multimodal}. For example, in a 24-layer LLM, we can loosely group layers into early, middle, and late stages: early layers typically encode local visual patterns (such as edges and textures) and basic token embeddings~\cite{tenney2019bert, zhang2025cross}; middle layers often capture cross-modal alignment~\cite{huang2025deciphering, yoon2025visual} (for example, region-to-phrase), object parts, relations, and shallow semantic groupings; and late layers are more associated with global scene understanding, abstract concept grounding, and high-level multimodal reasoning~\cite{rethinking2025visual, zhao2025accelerating, vtw2025boosting}. This motivates extracting and fusing representations from multiple depths rather than relying only on the last layer.

\noindent\textbf{LinkedOut representation definition.}
Since different layers encode different levels of visual and semantic knowledge, and the final text output is only a projection of the internal reasoning of a VLLM, our goal is to extract both old and new tokens across multiple layers so that we preserve and concentrate the world-knowledge embeddings inside the model \cite{survey2025mllm, storm2025token, rethinking2025visual}. We formalize \textit{LinkedOut} as a content-aware, knowledge-grounded representation that reuses a pretrained video LLM as a feature extractor and exposes world knowledge through prompts. Let a video $v$ be sampled into frames $\{x_t\}_{t=1}^{T}$, and let optional side channels (for example, ASR transcripts, metadata, tags) be $s$. A natural-language prompt $p$ steers the backbone toward task-relevant attributes. The video LLM $\mathcal{L}$ has $L$ transformer blocks and produces hidden states $\{H^{(\ell)}\}_{\ell=1}^{L}$ for the joint sequence of visual and textual tokens. Below we describe how these layer-wise token features are collected; the Cross-Layer Knowledge MoE Fuser is presented in Section~\ref{sec:moe}. Figure~\ref{fig:framework_moe} illustrates the two core modules: the Token Compressor Expert and the Cross-Layer  MoE.

\subsection{Raw World-Knowledge Token Extraction}
Because world knowledge is encoded in the pretrained transformer weights and applied to token embeddings at every layer, we directly extract token representations from intermediate layers of the LLM. Our Raw World-Knowledge Token Extraction module replaces predefined or handcrafted labels with dense intermediate token features obtained from a frozen (or selectively tuned) video LLM~\cite{alayrac2022flamingo, li2023blip2, liu2023llava}. Concretely, each frame $x_t$ is tokenized into patches by a vision tokenizer $g(\cdot)$, projected to the LLM embedding space by a lightweight adaptor $\phi$ (initialized from the vision–text bridge of the backbone), and concatenated with text tokens:
\begin{equation}
Z = \big[ \phi(g(x_1)), \ldots, \phi(g(x_T)), \ \mathrm{Tok}(p, s) \big].
\end{equation}
The backbone $\mathcal{L}$ then processes $Z$ and produces layer-wise hidden states $H^{(\ell)} \in \mathbb{R}^{N_\ell \times d}$, where $N_\ell$ is the token length at layer $\ell$ and $d$ is the model width.

\subsection{Cross-Layer Knowledge-Fusion MoE}
The Cross-Layer Knowledge-Fusion MoE forms the \textit{LinkedOut} representation from a VLLM in two steps. First, the \textit{Layer Token Compressor Expert} condenses old and new tokens within each layer into compact features. Second, the \textit{Cross-Layer Knowledge MoE Fuser} assigns data-dependent weights across layers and combines their features to produce a unified, knowledge-aware item embedding. These two components are tightly coupled by design; removing either breaks the construction of \textit{LinkedOut}. 




\subsubsection{Layer Token Compressor Expert}
Tokens at each layer can be numerous, creating a heavy computational burden and making it hard for downstream modules to determine each layer’s contribution. To make layer-wise features compact and comparable, we assign a token-compression expert to every layer, denoted $C^{(\ell)}$ (see the left panel of Figure~\ref{fig:framework_moe}). Each expert processes old and new tokens with separate branches, since old tokens mainly encode accumulated context while new tokens capture freshly generated knowledge. Concretely, the old branch first performs token merging, then compression; the new branch compresses tokens directly. We finally concatenate the two branch outputs to form the layer representation. 

\[
\mathbf{e}^{(\ell)} =
\mathrm{concat}\!\left(
C_{\text{old}}^{(\ell)}\!\big(\mathrm{Merge}_{\text{old}}^{(\ell)}(H_{\text{old}}^{(\ell)})\big),
\;
C_{\text{new}}^{(\ell)}\!\big(H_{\text{new}}^{(\ell)}\big)
\right).
\]
Here $C_{\text{old}}^{(\ell)}$ and $C_{\text{new}}^{(\ell)}$ are lightweight attention-pooling modules with learnable queries; the old branch uses more aggressive merging~\cite{bolyatoken} to reduce redundancy. We apply this to every $N$ layers (for example, every 2 or 4 layers), which empirically preserves the progression of world knowledge across depth while keeping the representation concentrated.

\subsubsection{Cross-Layer Knowledge MoE Fuser}\label{sec:moe}

Transformers at different depths in a VLLM organize information from local patterns to global abstractions. We have already concentrated knowledge in the old and new tokens of each layer, but we still need to learn which layers are most useful for a given downstream instance.

To select the right semantic granularity per video item, we fuse layer-wise features with a Mixture-of-Experts (MoE)~\cite{fedus2021switch, riquelme2021moe} (see the right panel of Figure~\ref{fig:framework_moe}). Each of the last $N$ tapped layers provides a compressed vector $\tilde{\mathbf{e}}^{(\ell)} \in \mathbb{R}^{d_c}$ (after token compression and prompt mixing). A per-layer expert then maps this vector to a common space:
\begin{equation}
    \mathbf{h}^{(\ell)} = E^{(\ell)}\big(\tilde{\mathbf{e}}^{(\ell)}\big) \in \mathbb{R}^{d_z},
\end{equation}
where $E^{(\ell)}$ is a lightweight MLP with residual connections. An item-conditioned gate then assigns soft weights over layers without circular dependence:
\begin{equation}
    \pi = \mathrm{Softmax}\!\left(G\big(\mathbf{z}_v\big)\right) \in \mathbb{R}^{N}, \quad
    \mathbf{z}_v = \sum_{\ell=1}^{N} \pi_\ell \, \mathbf{h}^{(\ell)},
\end{equation}
where $G(\cdot)$ is a gating MLP. The resulting embedding $\mathbf{z}_v$ is the unified, world-knowledge-aware item representation used by the ranker.
We use dense soft weights over all tapped layers. A sparse top-$k$ variant with load balancing can be adopted if desired, but is unnecessary here given the modest number of experts and layers.



\begin{table}[t]
\caption{Statistics of MicroLens-50K and MicroLens-100K.}
\label{tab:dataset_stats}
\vspace{-2mm}
\centering
\small
\setlength{\tabcolsep}{4pt}
\scalebox{0.85}{%
\begin{tabular}{@{}lcc@{}}
\toprule
\textbf{Statistics} & \textbf{MicroLens-50K} & \textbf{MicroLens-100K} \\
\midrule
\# Users & 50,000 & 100,000 \\
\# Items (Videos) & 19,220 & 19,738 \\
\# Total interactions & 359,708 & 719,405 \\
\# Training samples & 239,511 & 478,355 \\
\# Validation samples & 50,000 & 100,000 \\
\# Test samples & 50,000 & 100,000 \\
\# Fine-grained Tags & 15,580 & 15,580 \\
Avg seq length & 7.19 & 7.19 \\
Interactions per User & 5--15 & 5--15 \\
Video Duration & $<$ 400 seconds & $<$ 400 seconds\\
\bottomrule
\end{tabular}%
}
\vspace{-2mm}
\end{table}

\begin{table}[t]
\caption{Comparison on MicroLens-50K and MicroLens-100K, the only publicly available raw-video datasets for video recommendation to our knowledge.}
\label{tab:microlens-upright}
\centering
\small
\setlength{\tabcolsep}{3pt}
\resizebox{\linewidth}{!}{%
\begin{tabular}{@{}l *{4}{S} *{4}{S}@{}}
\toprule
& \multicolumn{4}{c}{\textbf{MicroLens-50K}} & \multicolumn{4}{c}{\textbf{MicroLens-100K}} \\
\cmidrule(lr){2-5} \cmidrule(l){6-9}
\textbf{Model} & \multicolumn{1}{c}{\makecell{\textbf{HR}\\\textbf{@10}}} & \multicolumn{1}{c}{\makecell{\textbf{HR}\\\textbf{@20}}} & \multicolumn{1}{c}{\makecell{\textbf{NDCG}\\\textbf{@10}}} & \multicolumn{1}{c}{\makecell{\textbf{NDCG}\\\textbf{@20}}} & \multicolumn{1}{c}{\makecell{\textbf{HR}\\\textbf{@10}}} & \multicolumn{1}{c}{\makecell{\textbf{HR}\\\textbf{@20}}} & \multicolumn{1}{c}{\makecell{\textbf{NDCG}\\\textbf{@10}}} & \multicolumn{1}{c}{\makecell{\textbf{NDCG}\\\textbf{@20}}} \\
\midrule
YouTube~\cite{covington2016deep}   & 0.0375 & 0.0632 & 0.0178 & 0.0245 & 0.0461 & 0.0747 & 0.0229 & 0.0301 \\
VBPR~\cite{he2016vbpr}     & 0.0544 & 0.0888 & 0.0273 & 0.0361 & 0.0624 & 0.1002 & 0.0314 & 0.0410 \\
MMGCN~\cite{wei2019mmgcn}     & 0.0403 & 0.0670 & 0.0197 & 0.0264 & 0.0214 & 0.0374 & 0.0103 & 0.0143 \\
LightGCN~\cite{he2020lightgcn}   & 0.0365 & 0.0534 & 0.0284 & 0.0345 & 0.0372 & 0.0618 & 0.0177 & 0.0239 \\
GRCN~\cite{wei2020graph}     & 0.0631 & 0.0982 & 0.0328 & 0.0415 & 0.0282 & 0.0497 & 0.0131 & 0.0185 \\
LayerGCN~\cite{zhou2023layer}  & 0.0627 & 0.0994 & 0.0320 & 0.0414 & 0.0730 & 0.1120 & 0.0382 & 0.0480 \\
BM3~\cite{zhou2023bootstrap}       & 0.0565 & 0.0918 & 0.0281 & 0.0372 & 0.0601 & 0.0975 & 0.0305 & 0.0401 \\
Freedom~\cite{zhou2023tale}   & 0.0656 & 0.1028 & 0.0334 & 0.0429 & 0.0654 & 0.1016 & 0.0337 & 0.0431 \\
MGCN~\cite{yu2023multi}      & 0.0708 & 0.1089 & 0.0363 & 0.0459 & 0.0717 & 0.1096 & 0.0371 & 0.0467 \\
MHCR~\cite{lyu2025multi}      & 0.0736 & 0.1102 & 0.0383 & 0.0477 & 0.0798 & 0.1187 & 0.0420 & 0.0519 \\
\midrule
\makecell[l]{\textbf{LinkedOut}\\\textbf{(Ours)}}
& \multicolumn{1}{c}{\textbf{0.0761}} & \multicolumn{1}{c}{\textbf{0.1127}} & \multicolumn{1}{c}{\textbf{0.0399}} & \multicolumn{1}{c}{\textbf{0.0491}} 
& \multicolumn{1}{c}{\textbf{0.1015}} & \multicolumn{1}{c}{\textbf{0.1477}} & \multicolumn{1}{c}{\textbf{0.0548}} & \multicolumn{1}{c}{\textbf{0.0664}}\\
\bottomrule
\end{tabular}%
}
\vspace{-4mm}
\end{table}

\subsection{Store-and-Retrieve Architecture}
We adopt a store-and-retrieve architecture for fast VLLM-based inference. As shown in Fig.~\ref{fig:framework_concept}(c–f), applying a VLLM directly at serving time is too slow. In practical recommendation systems, however, video feature extraction runs offline and is updated periodically, separate from the online inference path.
Since $\tilde{\mathbf{e}}^{(\ell)}$ are precomputed offline for each item (and for each prompt bank), the online stage only evaluates the gating network and a small set of selected experts. The gate weights $\pi$ indicate which abstraction levels contribute to the final score, and the prompt mixture reveals which attributes were emphasized, yielding human-readable explanations.

Offline item embeddings are obtained by running the extractor on all catalog videos with a small set of canonical prompts. The resulting per-layer vectors are stored in a feature store keyed by item IDs, which enables low-latency retrieval and online ranking without re-encoding videos. The extractor scales linearly with the number of frames $T$ and the number of selected layers $N$. Token compression reduces memory from $O(T N d)$ to $O(d_c)$ per item, where $d$ is the original token dimension and $d_c$ is the compressed world-knowledge dimension, which accelerates encoding and makes the store-and-retrieve pipeline practical at scale.

\section{Experiments}\label{sec_experiment}
\begin{table}[t]
\caption{Performance Comparison by Video-Recommendation Model Types on MicroLens-100k benchmark.}\label{tab:performance1}
\vspace{-1mm}
\label{tab:micolens100k}
\centering
\scriptsize
\setlength{\tabcolsep}{3pt}
\scalebox{1.0}{%
\begin{tabular}{@{}L{18mm}>{\centering\arraybackslash}p{1.15cm}SSSS@{}}
\toprule
\textbf{Model} & \textbf{Type} & \multicolumn{1}{c}{\makecell{\textbf{HR}\\\textbf{@10}}} & \multicolumn{1}{c}{\makecell{\textbf{NDCG}\\\textbf{@10}}} & \multicolumn{1}{c}{\makecell{\textbf{HR}\\\textbf{@20}}} & \multicolumn{1}{c}{\makecell{\textbf{NDCG}\\\textbf{@20}}} \\
\midrule
DSSM~\cite{huang2013dssm}      & \multirow{4}{*}{\makecell{IDRec\\(CF)}} & 0.0394 & 0.0193 & 0.0654 & 0.0258 \\
LightGCN~\cite{he2020lightgcn} &  & 0.0372 & 0.0177 & 0.0618 & 0.0239 \\
NFM~\cite{he2017nfm}           &  & 0.0313 & 0.0159 & 0.0480 & 0.0201 \\
DeepFM~\cite{guo2017deepfm}    &  & 0.0350 & 0.0170 & 0.0571 & 0.0225 \\
\midrule
NextItNet~\cite{yuan2019nextitnet} & \multirow{3}{*}{\makecell{IDRec\\(SR)}} & 0.0805 & 0.0442 & 0.1175 & 0.0535 \\
GRU4Rec~\cite{hidasi2016gru4rec}   &  & 0.0782 & 0.0423 & 0.1147 & 0.0515 \\
SASRec~\cite{kang2018sasrec}       &  & 0.0909 & 0.0517 & 0.1278 & 0.0610 \\
\midrule
YouTubeID~\cite{covington2016youtube} & \multirow{8}{*}{\makecell{VID\\Rec}} & 0.0461 & 0.0229 & 0.0747 & 0.0301 \\
YouTubeID+V~\cite{covington2016youtube}  &  & 0.0392 & 0.0188 & 0.0648 & 0.0252 \\
MMGCNID~\cite{MMGCN}                     &  & 0.0141 & 0.0065 & 0.0247 & 0.0092 \\
MMGCNID+V~\cite{MMGCN}                   &  & 0.0214 & 0.0103 & 0.0374 & 0.0143 \\
GRCNID~\cite{wei2020graph}               &  & 0.0282 & 0.0131 & 0.0497 & 0.0185 \\
GRCNID+V~\cite{wei2020graph}             &  & 0.0306 & 0.0144 & 0.0547 & 0.0204 \\
DSSMID+V~\cite{huang2013dssm}            &  & 0.0279 & 0.0137 & 0.0461 & 0.0183 \\
SASRecID+V~\cite{kang2018sasrec}         &  & 0.0799 & 0.0415 & 0.1217 & 0.0520 \\
\midrule
NextItNetV~\cite{yuan2019nextitnet}      & \multirow{3}{*}{\makecell{Video\\Rec}} & 0.0862 & 0.0468 & 0.1246 & 0.0562 \\
GRU4RecV~\cite{hidasi2016gru4rec}        &  & 0.0954 & 0.0517 & 0.1377 & 0.0623 \\
SASRecV~\cite{kang2018sasrec}            &  & 0.0948 & 0.0515 & 0.1364 & 0.0619 \\
\midrule
\makecell[l]{\textbf{LinkedOut}\\\textbf{(Ours)}} & \makecell{Video\\LLMRec} & \textbf{0.1015} & \textbf{0.0548} & \textbf{0.1477} & \textbf{0.0664} \\
\bottomrule
\end{tabular}
}
\vspace{-2mm}
\end{table}

\subsection{Datasets}

We evaluate on MicroLens-50K and MicroLens-100K~\cite{ni2023microlens}, two large-scale, content-driven micro-video recommendation benchmarks that, to the best of our knowledge, are the \textit{only} public datasets providing raw videos for content-aware modeling. Table~\ref{tab:dataset_stats} summarizes key statistics. Each video includes rich multimodal content, such as raw frames (mostly under 400 seconds), audio tracks, titles, cover images, and user comments, which makes these datasets suitable for evaluating foundation-model-driven recommendation. Interactions are timestamped comments spanning 15{,}580 fine-grained tags (for example, food, sports, travel, entertainment), enabling sequential modeling. We follow the official training and testing protocol in~\cite{ni2023microlens}.



\subsection{Benchmarks}
We compare LinkedOut with state-of-the-art video recommendation baselines, which can be broadly categorized into three groups:
(1) \textbf{IDRec} \cite{ni2023microlens} methods rely solely on item IDs without leveraging video content. They include collaborative filtering (CF) approaches: \textbf{DSSM} \cite{huang2013dssm} learns user/item embeddings via dual-tower matching; \textbf{LightGCN} \cite{he2020lightgcn} applies graph convolution on the user-item bipartite graph; \textbf{NFM} \cite{he2017nfm} models second-order feature interactions with neural bi-interaction; \textbf{DeepFM} \cite{guo2017deepfm} combines factorization machines with deep networks. Sequential recommendation (SR) models include \textbf{GRU4Rec} \cite{hidasi2016gru4rec} with recurrent units, \textbf{SASRec} \cite{kang2018sasrec} with self-attention, and \textbf{NextItNet} \cite{yuan2019nextitnet} with dilated convolutions.
(2) \textbf{VIDRec} methods augment IDRec with frozen, pre-extracted video features as side information, denoted by the ``+V'' suffix (e.g., \textbf{YouTubeID+V} \cite{covington2016youtube}, \textbf{MMGCNID+V} \cite{MMGCN}, \textbf{GRCNID+V} \cite{wei2020graph}, \textbf{DSSMID+V}, \textbf{SASRecID+V}). These methods concatenate video embeddings with ID embeddings while keeping the video encoder frozen. \textbf{YouTubeID} \cite{covington2016youtube} is a two-stage system with candidate generation and ranking; \textbf{MMGCN} \cite{MMGCN} propagates user-item signals separately per modality; \textbf{GRCN} \cite{wei2020graph} refines collaborative filtering with graph-based representations.
(3) \textbf{VideoRec} \cite{ni2023microlens} methods jointly train a video encoder with the recommender via end-to-end (E2E) learning, replacing ID embeddings with learnable video representations (\textbf{GRU4RecV} \cite{hidasi2016gru4rec}, \textbf{SASRecV} \cite{kang2018sasrec}, \textbf{NextItNetV} \cite{yuan2019nextitnet}). These E2E models are computationally expensive but achieve superior performance by optimizing video understanding and recommendation simultaneously \cite{ni2023microlens}.

\begin{table}[t]
  \caption{Ablation study of component effectiveness analysis.}\label{tab:ablation}
  \vspace{-2mm}
  \centering
  \footnotesize
  \resizebox{1.0\linewidth}{!}{%
  \begin{tabular}{@{}lcccc@{}}
    \toprule
    \textbf{Variant} & \textbf{HR@10} & \textbf{NDCG@10} & \textbf{HR@20} & \textbf{NDCG@20} \\
    \midrule
    Last layer last token & 0.0763 & 0.0404 & 0.1121 & 0.0494 \\
    Mean pooling+MoE & 0.0888 & 0.0471 & 0.1291 & 0.0572 \\
    Last token+MoE & 0.0958 & 0.0518 & 0.1394 & 0.0628 \\
    \midrule
    \textbf{LinkedOut (full)} & \textbf{0.1015} & \textbf{0.0548} & \textbf{0.1477} & \textbf{0.0664} \\
    \bottomrule
  \end{tabular}%
  }
  \vspace{-3mm}
\end{table}

\subsection{Implementation Details}

\textbf{VLLM extraction.}
We use LLaVA-OneVision~\cite{li2024llava} 7B as the VLLM backbone and freeze its parameters during feature extraction. We attach PyTorch hooks to capture token embeddings during generation: as the model decodes, previously generated tokens are first processed by the decoder and new tokens are produced one by one. To reduce computation, we record token representations every four transformer layers inside the VLLM. 

\noindent\textbf{Recommendation model training.}
We train on 8$\times$ NVIDIA H100 GPUs with mixed precision and a batch size of 256 sequences. Each user history contains at most 10 videos. Unless noted otherwise, we train for 100 epochs with AdamW and parameter-group-specific learning rates: 0.0001 for the video encoder (when unfrozen), 0.0001 for the text and image encoders (with pooler and classifier heads separated), and 0.00001 for the recommendation head. We set weight decay to \(0.1\) and clip the global gradient norm at \(5\). Images are resized to \(224\times224\).
We use a \texttt{step\_schedule\_with\_warmup} scheduler at the epoch granularity (gap \(=1\) epoch, \(\alpha=1.0\)) with zero warmup steps by default. The total loss combines an alignment term and a uniformity term, reported per batch, and is optimized jointly with the recommendation objective.

\subsection{Performance Analysis}
\textbf{Performance Comparison by Video-Recommendation Model Class}. 
Table~\ref{tab:micolens100k} compares LinkedOut with state-of-the-art baselines on MicroLens-100K across four categories. LinkedOut achieves the best performance with HR@10 of 0.1015 and NDCG@10 of 0.0548, outperforming all baseline approaches by substantial margins. Compared to the strongest ID-based sequential recommender (SASRec, HR@10: 0.0909), LinkedOut achieves an 11.7\% relative improvement, demonstrating that foundation-model-driven video understanding provides richer semantics beyond collaborative filtering patterns. Against the best end-to-end VideoRec model (GRU4RecV, HR@10: 0.0954), LinkedOut achieves a 6.4\% relative gain, validating that our layer-wise MoE fusion and token compression effectively extract more discriminative features than standard video encoders. Notably, most VIDRec methods (e.g., MMGCNID+V, GRCNID+V) underperform their ID-only counterparts, suggesting that naively concatenating frozen video features with ID embeddings introduces noise rather than useful signals. In contrast, LinkedOut's adaptive layer selection and token-level aggregation enable fine-grained control over which semantic levels contribute to recommendation, avoiding the feature degradation observed in VIDRec approaches. The consistent improvements across all metrics (HR@10, NDCG@10, HR@20, NDCG@20) confirm that LinkedOut captures both relevance and ranking quality more effectively.

\begin{figure}[t]
\centering
\includegraphics[width=0.85\linewidth, height=0.35\linewidth]{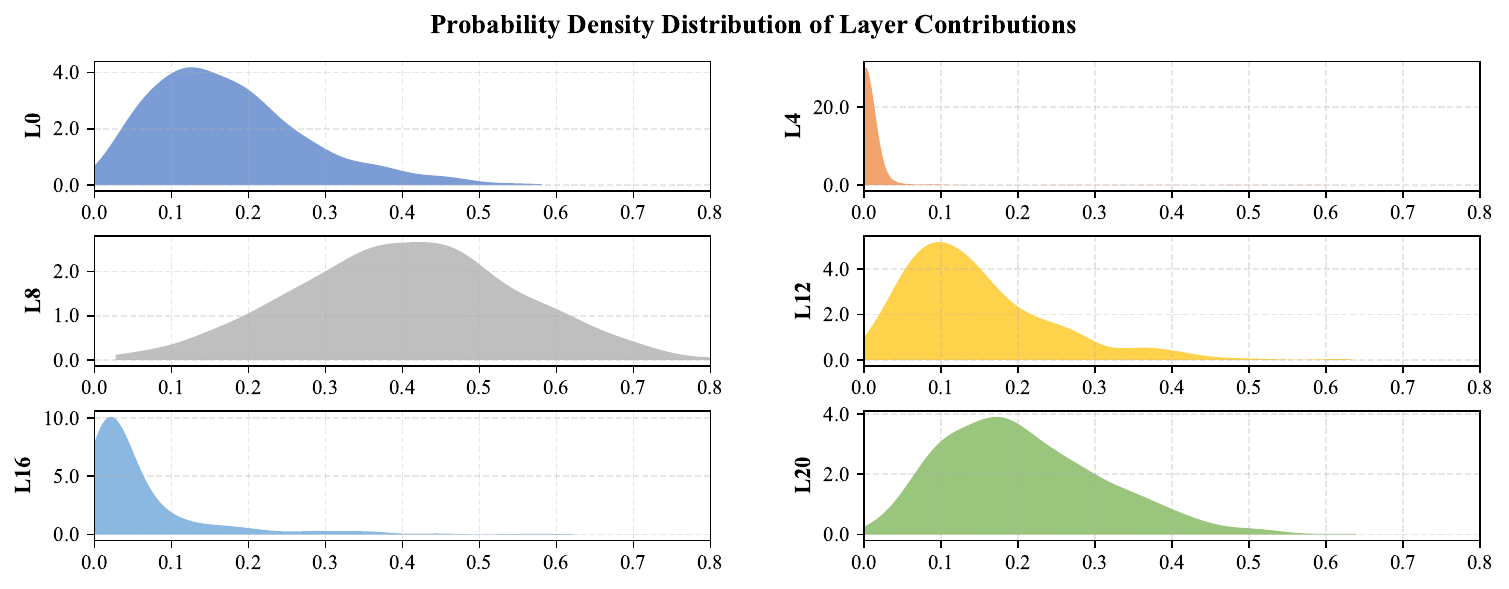}
\vspace{-3mm}
\caption{Probability density distribution of layer-wise MoE gate weights via kernel density estimation (KDE). Each subplot shows the distribution of normalized contribution values for a transformer layer (L0, L4, L8, L12, L16, L20). Distribution width and peak location reveal utilization patterns, where L8 shows the broadest distribution centered at 0.4, while L4 and L16 exhibit more concentrated distributions with relatively smaller contributions.}
\label{fig:moe_kde}
\vspace{-3mm}
\end{figure}

\noindent\textbf{Comparison across datasets.}
Table~\ref{tab:microlens-upright} compares \textit{LinkedOut} with recent baselines on the two public raw-video datasets, MicroLens-50K and MicroLens-100K. \textit{LinkedOut} improves all four metrics (HR@10, HR@20, NDCG@10, NDCG@20) on both datasets, indicating consistent performance across different dataset sizes and content complexity.

\subsection{Ablation Study}
\textbf{Component effectiveness analysis.}
To validate the importance of our design choices, we conduct ablation studies on the components. Table~\ref{tab:ablation} reports results on MicroLens-100K. First, we evaluate a baseline that uses only the last layer last token without MoE fusion. This simpler approach mirrors standard LLM extraction practices, resulting in significant performance drops (HR@10: 0.0763 vs. 0.1015), demonstrating that both layer-wise fusion and token-level aggregation are crucial. Second, we examine mean pooling with MoE, which applies layer-wise fusion but uses simple average pooling across all tokens at each layer. While this variant benefits from multi-layer knowledge (HR@10: 0.0888), it treats all tokens equally and fails to emphasize recommendation-relevant features. Third, we test last token with MoE, where we apply layer-wise fusion but only use the final token from each layer. This variant achieves better performance (HR@10: 0.0958), suggesting that the last token contains more task-relevant information, yet it still underperforms our complete model by ignoring rich intermediate token representations. These ablations confirm that LinkedOut's complete design is essential for achieving SOTA performance. Each component contributes meaningfully to the final results.

\begin{figure}[t]
\centering
\includegraphics[width=0.9\linewidth, height=0.40\linewidth]{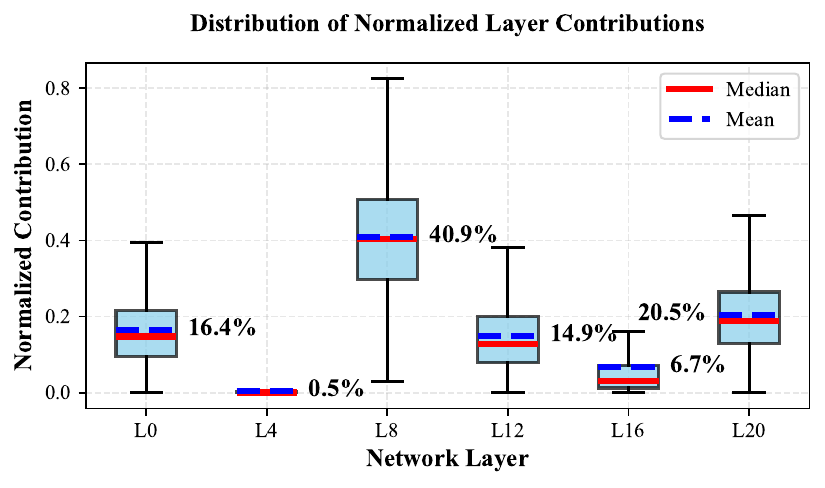}
\vspace{-10pt}
\caption{Statistical summary of layer-wise MoE gate contributions. Box plots show distribution of normalized contribution values per layer, with red lines (median) and blue dashed lines (mean). Annotated percentages are average contributions, where L8 dominates at 40.9\%, followed by L20 (20.5\%) and L0 (16.4\%), while L4 (0.5\%) and L16 (6.7\%) contribute relatively less.}
\label{fig:moe_boxplot}
\vspace{-15pt}
\end{figure}

\noindent\textbf{Layer-wise contribution analysis.}
To understand which semantic levels are most valuable for video recommendation, we analyze the learned MoE gate weights across all validation samples. We visualize layer contributions through two complementary perspectives: Figure~\ref{fig:moe_kde} shows probability density distributions via kernel density estimation (KDE) for each tapped layer (L0, L4, L8, L12, L16, L20); Figure~\ref{fig:moe_boxplot} provides box plot statistics of median, mean, and distribution spread, annotated with average contribution percentages. Our key finding is that intermediate layers, particularly L8, dominate with 40.9\% contribution, significantly outperforming early layers (L0: 16.4\%) and later layers (L20: 20.5\%), while L4 and L16 contribute minimally (0.5\% and 6.7\%). The KDE distributions further show that L8 exhibits broad, stable patterns across diverse videos, whereas L4 and L16 are more concentrated. This confirms that video recommendation benefits from adaptive multi-level semantic features, with mid-level representations balancing visual details and abstract concepts rather than relying solely on the deepest layers as commonly assumed in vision-language models.

\begin{table}[t]
\caption{Inference-time comparison for potential design of \emph{VLLM-based} recommendation. We report latency for direct VLLM serving and for our Store-and-Retrieve design; a traditional non-VLLM module is shown only as a reference for systems-level latency and is not a semantic baseline. The goal is to bring VLLM-based approaches to practical serving latency.}
\label{tab:time-comp}
\centering
\small
\setlength{\tabcolsep}{5pt}
\resizebox{1.0\linewidth}{!}{%
\begin{tabular}{@{}c l l l@{}}
\toprule
\textbf{VLLM-based?} & \textbf{Description} & \textbf{Type} & \textbf{Time} \\
\midrule
\xmark & Traditional video recommendation \textit{[reference only]} & Fig.\ref{fig:framework_concept} (a,b) & 0.864 ms \\
\midrule
\cmark & Direct VLLM-based & Fig.\ref{fig:framework_concept} (c,d) & 5510 ms \\
\cmark & \makecell[l]{Store-and-Retrieve VLLM + Video \\ recommendation module (LinkedOut)} & Fig.\ref{fig:framework_concept} (f) & 5.964 ms \\
\bottomrule
\end{tabular}%
}
\vspace{-2mm}
\end{table}

\noindent\textbf{Efficiency Analysis.}
To demonstrate the computational advantages of our Store-and-Retrieve architecture, we compare inference latency across three approaches on a single NVIDIA H100 GPU (Table\ref{tab:time-comp}). Traditional video recommendation without VLLM (Fig.\ref{fig:framework_concept} a,b) achieves 0.864ms per inference, while direct real-time VLLM inference (Fig.\ref{fig:framework_concept} c,d) incurs a prohibitive 5.51s latency, making it infeasible for deployment. LinkedOut's Store-and-Retrieve paradigm (Fig.\ref{fig:framework_concept} f) decouples offline feature extraction (5.02s per video when batch-processing 100 videos, paid once) from online inference, where MoE fusion (5.1ms) and recommendation ranking (0.846ms) total just 5.964ms per query nearly 1000× faster than direct VLLM while preserving rich semantic understanding. This efficiency gain validates that foundation-model-driven video recommendation is practical at scale.

\section{Conclusion}\label{sec_conclusion}
We presented LinkedOut, a knowledge-aware video recommendation framework that links world knowledge out of raw pixels via pretrained video LLMs. By extracting semantically grounded token representations directly from frames, LinkedOut eliminates the language bottleneck and hand-crafted-label constraints of conventional approaches. Our promptable feature focus module enables flexible steering without retraining, while the cross-layer MoE adaptively fuses representations across transformer depths to select the right semantic granularity per instance. Experiments on MicroLens-50K and MicroLens-100K demonstrate state-of-the-art performance, with consistent gains in personalization, cold-start, and long-tail scenarios. Ablation studies confirm the importance of layer-wise fusion, promptable extraction, and adaptive semantic depth selection. These results highlight a practical path toward next-generation recommendation systems that fully leverage video LLM world knowledge and pixel-level visual semantics for robust, scalable video recommendation.

{
    \small
    \bibliographystyle{ieeenat_fullname}
    \bibliography{main}
}


\end{document}